# Moving Object Detection for Event-based Vision using k-means Clustering

Anindya Mondal , Mayukhmali Das

1Department of Electronics and Telecommunication Engineering

Jadavpur University, Kolkata, India

{anindyam.jan, mayukhmalidas322}@gmail.com

## Abstract

*Moving object detection is a crucial task in computer vision. Event-based cameras are bio-inspired cameras that mimic the working of the human eye. Unlike conventional frame-based cameras, these cameras pose multiple advantages, like reduced latency, HDR, reduced motion blur during high motion, low power consumption, etc. However, these advantages come at a high cost, as event-based cameras are sensitive to noise and have low resolution. Moreover, for the lack of useful visual features like texture and colour, moving object detection in these cameras becomes more challenging. Our proposed method uses k-Means clustering for detecting moving objects in event-based data. We further compare the proposed method against state-of-the-art algorithms and show performance improvement over them.*

***Keywords*:** Event-based cameras, Moving Object Detection, k-Means Clustering, Silhouette Analysis.

## 1: Introduction

Event-based cameras are bio-inspired cameras [1]. While frame-based cameras capture images at a definite frame rate which is determined by an external clock, each pixel in event-based cameras memorizes the log intensity each time an event is sent and simultaneously monitors for a sufficient change in magnitude from this memorized threshold value [1]. The event is recorded by the camera and is transmitted by the sensor in the form of its location {x, y}, its time of occurrence (timestamp) t and its polarity p (taking a binary value 1 or −1, representing whether the pixel is brighter or darker) [2]. The working of an event-based camera is shown in Fig. 1. The sensors used in event-based cameras are data-driven, for their output depends on the amount of motion or brightness change in the scene [1]. Higher is the motion, higher is the number of events generated. The events are recorded in microsecond resolution and are transmitted in sub-millisecond latency, making these sensors react quickly to visual stimuli [1]. Thus, while frame-based cameras capture the absolute brightness of a scene, event-based cameras capture the per-pixel brightness asynchronously, making traditional computer vision algorithms inapplicable to be implemented for processing the event data. Detection of moving objects is an important task in automation, where a computer separates a moving object from a stationary one. Through the years, various algorithms have been designed for detecting moving objects in a scene [3]. However only a very few of them are applicable for event-based data. One of the many reasons includes the lack of sufficient visual features like colour and texture in event data, as event-based sensors only capture the relative changes in brightness of a scene. Some other reasons include lack of sufficient training data (restricting the application of deep learning methods), the abundance of noise (generated due to sensor defects for the technological novelty of event-based sensors), etc. However, there have been multiple attempts to address the problem of moving object detection and/ or tracking in event-based data, like in [4] and [5], where they have used parametric models and in [6]–[8], where they have used traditional clustering algorithms. But using parametric models require the optimization of multiple parameters. Also, traditional algorithms require extensive pre-processing of the event data and their success depends upon the careful choice of the affecting parameters [7]. In recent years the use of graphs to tackle computer vision problems has increased considerably [9], [10]. This is because graphs can represent data lying on irregular and complex domains much better than other data structures [11]. Now, in event-based cameras, the neuromorphic sensors activate in time in an asynchronous manner. So the event data streams are produced at irregular space-time coordinates, depending upon the scene activity [12]. Thus, by using the graph-based method, we maintain the asynchronicity and sparsity of the event data and utilize the advantages that come out of it [13]. To summarize our contributions, we are the first to introduce k-means clustering [15] for detecting moving objects in event-based data. Experimental results show that our method improves upon the previous clustering-based approaches to detect moving objects in event-based data. Along with that, we automate the process of determining the optimal number of clusters by using silhouette analysis [16]. The paper is organized as follows: In Section 2 we discuss the related works. We discuss the methodology in Section 3. In Section 4 we discuss the experiments and results. Finally, in Section 5 we discuss the conclusion and scope for future work.

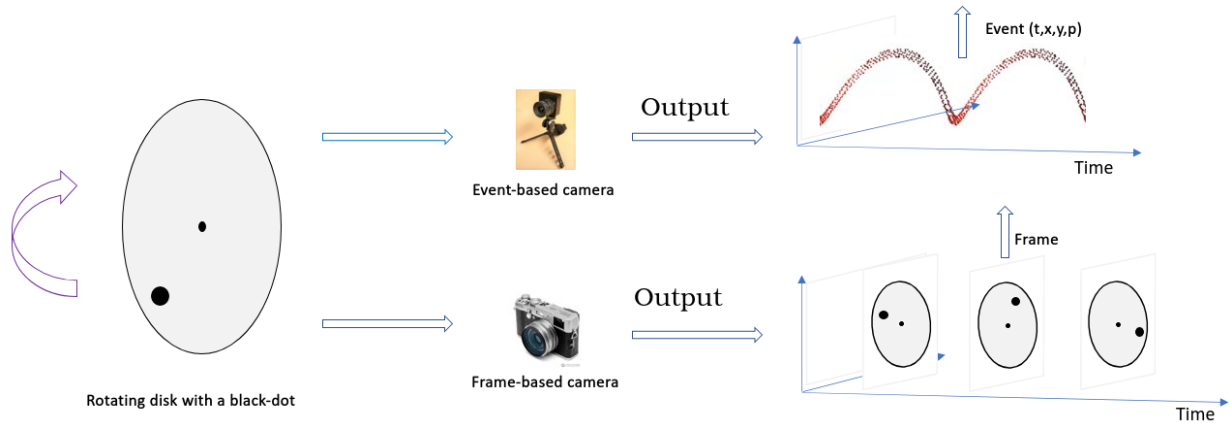

Fig. 1: Visualization of the output from an event-based camera and a standard frame-based camera while they face a rotating disk with a black dot. Inspired by [1]

## 2: Related Works

In this section, we will discuss some of the available frameworks for moving object detection and we also cover some of its extensions to event-based vision. The problem of moving object detection has been widely studied in computer vision [32]. Some of the approaches include the use of deep learning [17], [18], geometrical understanding of a scene [19], etc. However, these methods have multiple shortcomings, especially when we try to extend them to event-based vision. For example, using deep learning techniques requires an enormous amount of labelled training data to evade over-fitting, which is unavailable for event-based data. Furthermore, event-based data lie on an irregular domain, hence geometrical understanding of a scene is not possible, as the concept of frames ceases to exist here. Now we address the requirement for developing moving object detection algorithms for event-based data. We know that conventional frame-based cameras suffer from high-motion blur during high-speed motion. Also, they (frame-based cameras) capture a considerable amount of redundant data, as they capture snapshots of a scene at regular intervals, thus requiring an enormous amount of power and data-processing facility. Event-based cameras significantly resolve these problems [1], [2], [20]. This is because event-based sensors capture the scene dynamics only (binary change in brightness occurs when motion occurs), so there is on-chip moving object detection from the static background [6]. So the task of detecting moving objects may seem trivial in event-based cameras. However, we must remember that, unlike their frame-based counterparts, event-based sensors capture only binary variations in brightness [6]. So events do not contain much visual information (like texture, colour, etc.) that exists in image frames [6], making the detection task more challenging for event-based cameras. There have been few attempts to perform moving object detection in event-based cameras by utilizing clustering-based strategies. In 2012, Piatkowska et.al. [6] have investigated the potential of Gaussian Mixture Models (GMM) [21] for multiple persons tracking in these cameras. In 2018, Chen et.al. [7] and Hinz et.al. [8] have performed multi-vehicle detection and tracking using classical clustering approaches (like DBSCAN [22], MeanShift [23], WaveCluster, etc. However, GMMs are noise sensitive, as they assume each data point (here the events) to be independent of its neighbours, thus neglecting the similarity relations among those points [24]. Furthermore, density-based approaches like DBSCAN are also ill-suited for dealing with sparse event-based data, as the data points (here events) are of varying density. Approaches like MeanShift involve optimization of multiple parameters [25], making the task of finding the most optimal value difficult. In our work, we implement k-means clustering for detecting the moving objects in event data. Using k-means improves upon the previous methods in various ways:

- Optimizing the number of clusters in k-means is sufficient, contrary to the previously discovered methodologies that needed joint optimization of multiple parameters.
- We automate the mode of determining the optimal value of the number of clusters by using silhouette analysis [16], thus decreasing the task of finding it (number of clusters) for each sequence manually.

## 3: Methodology

Our work involves the application of k-means clustering [15], [26] for detecting moving objects in event-based data. As the event points are sparse and spatiotemporally separated from one another, k-means proves to be a suitable method for this detection task. The schematic diagram of our work is shown in Fig. 2. Firstly we have taken a sampling strategy to select a small set of events. This is done to make the processing of event data computationally efficient. Then we construct a graph using those selected event points by following a k-nearest neighbourhood (k-NN) strategy. Following that we apply k-means clustering on the k-NN graph. Finally, we find the optimum value of the number of clusters in K-means clustering by performing silhouette analysis on those individual clusters. These clusters here represent the moving objects.

   A. **Sampling:** In the event-based vision, the events generated as a result of object motion can be treated as point-cloud data in 3D spacetime volume [2]. Let us consider there are N events and their corresponding M image frames captured at uniform timestamp $T_i$ (i ranges from 1 to M). For generating meaningful samples, we divide the whole event space-time volume into M partitions.

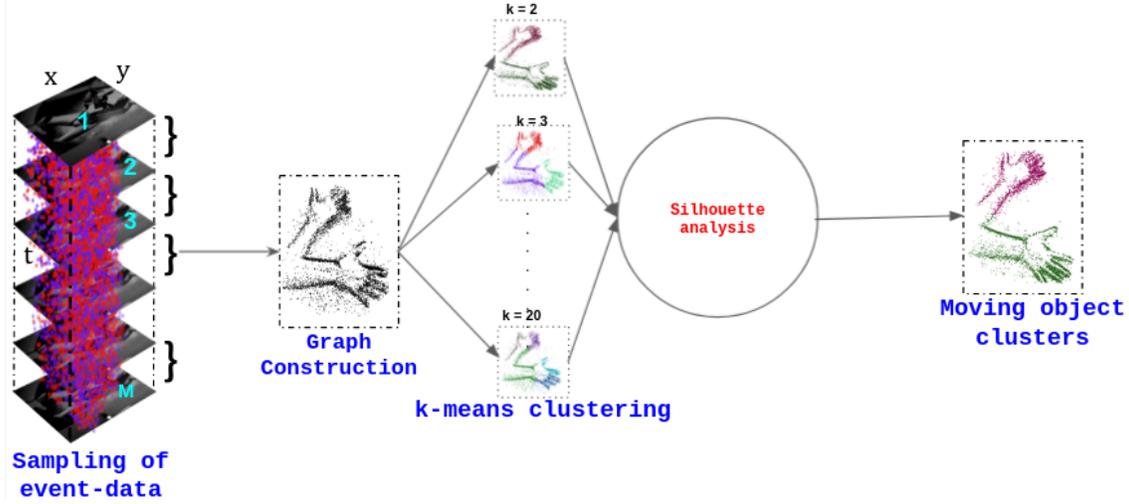

Fig. 2: Overview of our proposed method

Let each partition contains Qi events that occurred before timestamp $T_i$. Thus $\sum_{i=1}^{M} Q_i = M$ Now we adopt a uniform sampling strategy for obtaining a small set of events. Let S events be selected from each of these partitions. Each event-point in S is represented as a tuple sequence:

$$\{e_i\}_S = \{x_i, y_i, t_i\}_S \quad \ldots\ldots(1)$$

where $\{x_i, y_i\}$ indicates the spatial address at which the spike event had occurred, ti is the timestamp indicating when the event was generated and S represents the total number of events.

B. **Graph construction and clustering:** Now we know that the asynchronously generated events are sparse in the Spatio-temporal domain (image plane evolving in time) and the underlying graph generated is grossly unstructured. This is opposite to what we get in the case of frame-based cameras, where the graph of pixels of an image is mostly regular [5]. So, to maintain this asynchronicity, we connect these S events into a Spatio-temporal graph. The graph is constructed by using a k-nearest-neighbours (kNN) strategy [27]. The neighbourhood is based on the spatiotemporal similarity between the events in the 3D point cloud. Let us define a graph G = G(V, E), where V is the set of nodes and E is the set of edges connecting them. Here we denote each event $e_i$ = $e_i(x_i, y_i, t_i)$ as a node $v_i$ in the graph ($v_i \in V$). We connect $v_i$ and $v_j$ with an edge $\varepsilon_{ij}$ ($\varepsilon_{ij} \in E$) if either $v_i$ is among the k nearest neighbours of $v_j$ or $v_j$ is among the k nearest neighbours of $v_i$. Here the value of k in k-NN has been taken by considering multiple issues. Firstly a small value of k may be detrimental to the performance of our model because it will make our model noise sensitive. Secondly, a higher value of k will make our model computationally expensive. For these reasons, we manually set the value of k as a trade-off between noise sensitivity and computational complexity. Now as we have the graphs of events, we can apply K-means clustering on each of these event graphs. In our work, the connected components (clusters) in the graph represents the moving objects. Let us consider there are f moving objects in a specific sequence (graph of a set of events in between two consecutive timestamps). This implies that there are f distinct clusters (f ≤ S). These f clusters are obtained by applying K-means clustering on the graph of S events. Let the clusters be $C_1, C_2, \cdots, C_f$. Each of those clusters represents the moving objects for a definite set of events.

## C. Finding the optimal number of clusters:

Finding the optimal value of the number of clusters (here f) in k-means clustering is a crucial task. Multiple methods have been proposed to determine its value, like Rand Index [28], Elbow method [26], Silhouette analysis [16], etc. Here we use Silhouette analysis to determine the clusters within the event points, which, in turn, represent the moving objects. In our work, we feed the model with some values of f (number of clusters), and the model performs silhouette analysis on those clusters for each value of f, giving the best possible value as the output. As mentioned in Section 2, using this method significantly reduces the need for supervision to find the best possible value of f for each sequence. Silhouette value measures the difference between the within-cluster tightness and separation from the rest [29]. It is defined for each data point and is composed of two scores:

- a(i): The mean distance between a data point i and all other points in the same cluster.
- b(i): The mean distance between a data point i and all other points in all other clusters of which it is not a member.

Now the silhouette value s(i) for that single point is given as:

$$s(i) = \frac{b(i) - a(i)}{max(a(i), b(i))} \quad .....(2)$$

Kauffman et. al. [30] proposed the term Silhouette Coefficient SC, which returns the maximum value of the mean s(i) over all data points of a specific sequence. SC is defined as:

$$SC = max_f \tilde{s}(f) \quad ......(3)$$

where $\tilde{s}(f)$ represents the mean s(i) over the entire data of a specific sequence for a definite number of clusters f. The value of SC is bounded between −1 and 1, where a score near 1 shows greater intra-cluster tightness and greater inter-cluster distance. A higher value means that the clusters are dense and well separated, leading to a meaningful representation of the clusters. In our work, we use SC to determine the optimal value of f. We take some values of f, ranging from 2 to 20 and plot their respective SCs. From those plots, we take the value of f for which the magnitude of SC is maximum. This f is the required optimal number of clusters. This is how we get the actual number of moving objects in a scene.

| Metrics | Formula |
|---|---|
| True positive (TP) | $\frac{|E \cap GT|}{|E \cup GT|} \geq 0.85$ |
| False positive (FP) | $\frac{|E \cap GT|}{|E \cup GT|} < 0.85$ |
| False negative (FN) | No ground truth detected |
| Precision (P) | $\frac{TP}{TP+FP}$ |
| Recall | $\frac{TP}{TP+FN}$ |
| F measure | $\frac{2 \times P \times R}{P+R}$ |

TABLE I: Evaluation metrics

# 4: Experiments and Results

In this section, we briefly discuss the implementation of our proposed method. First, we mention the dataset used and discuss the metrics in Section 4. A and 4. B respectively. In Section 4. B, we also compare our method against the state-of-the-art baselines [6]–[8] and show the improvements we have made over them.

| Sequence | DBSCAN | | | Meanshift | | | GMM | | | k-Means | | |
|---|---|---|---|---|---|---|---|---|---|---|---|---|
| | Recall | Precision | F measure | Recall | Precision | F measure | Recall | Precision | F measure | Recall | Precision | F measure |
| Hands | 66.57 | 77.33 | 71.55 | 71.56 | 79.30 | 75.23 | 87.14 | 88.67 | 87.90 | **89.41** | **88.43** | **88.92** |
| Cars | 48.58 | 31.50 | 38.22 | 41.68 | 50.00 | 45.46 | 50.26 | 77.60 | 61.00 | **53.71** | **79.23** | **64.02** |

TABLE II: Quantitative comparison results. Note that the precision and recall scores are in percentage. The best results are in bold.

Some fail cases have been discussed too in Section 4. B. As mentioned in Section 3. B, we have customized the value of k in k-NN through experiments. Here we set the value of nearest neighbours at 45 for the hand's sequence and 200 for the cars sequence of the dataset. All the experiments were performed on an Intel® Core i5 8th Gen. PC, having 4 CPUs with a clock frequency of 4 GHz each.

A. **Dataset:** We evaluate our model on the publicly available DVSMOTION20 [31] dataset. The events here are recorded using the IniVation DAViS346 camera. This camera has a 346×260 spatial resolution and outputs frames up to 60 frames per second for RGB/ grayscale images and in microsecond resolution for the events. This dataset contains sequences (both events and their corresponding grayscale images) captured in both indoor and outdoor environments, involving multiple moving objects.

B. **Metrics and Comparison:** As moving object detection (MOD) is a widely discussed problem in the computer vision regimen, various evaluation metrics have been proposed over the years. However, for the novelty of event-based vision technology, not many of them can be applied for the task of MOD here. So we follow the method prescribed in [7], where they have accumulated the events corresponding to their grayscale frames in different time intervals. We treat the grayscale images in the dataset as the ground truth and perform a coverage test [6] on them with the detected objects in event-based data. We have manually drawn bounding boxes covering the moving objects in the ground truth and the testing data (clusters of events). Now we calculate the percentage overlap of the area of bounding boxes in the ground truth (denoted by GT) and the test data (denoted by E) to check the performance of our model. More details on the metrics can be found in Table I.

The visual results and comparison with the SOTA models can be found in Fig. 3. Quantitative comparisons can be found in Table II. Visual results clearly depict the improvement in performance for detecting moving objects in event-based data over the previous attempts. Notable improvements can be found during occlusions and with the increase in the number of moving objects in a scene. However, there are some cases where our model has performed poorly. Fig.4 shows some of such fail cases. Significant deviation from the expected results can be seen during the abundance of noise in the event data. Poor performance can also be detected when one of the moving objects is significantly larger than the others, where using k-means have split that large object into multiple clusters.

# 5: Conclusion and Future Works

In this paper, we demonstrate the performance of k-means clustering for detecting moving objects in event-based data. We show that using k-means significantly improves upon the previous approaches. We also propose a method that has reduced the necessity for supervision for determining the optimal number of clusters in a sequence. We have also shown our method's flexibility and versatility by evaluating our way on a dataset covering both indoor and outdoor scenes. This work will be beneficial for multiple tasks in computer vision, robotics, self-driving cars, etc. Future works include the addition of motion models for tracking moving objects alongside detecting them. Also combining various clustering algorithms for the detection task can be a future research direction.

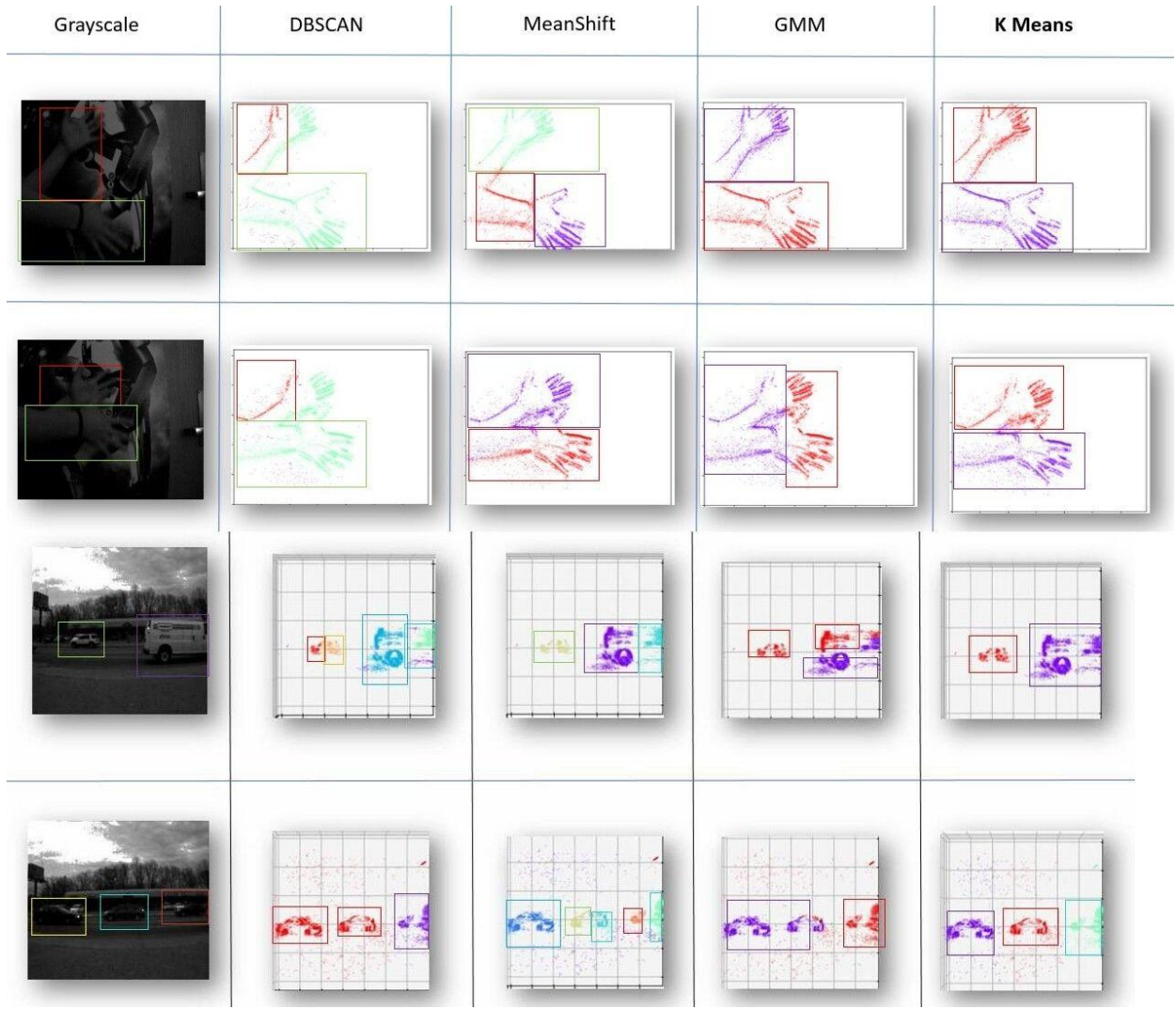

Fig. 3: Visual comparison with the SOTA methods (Best viewed in colour)

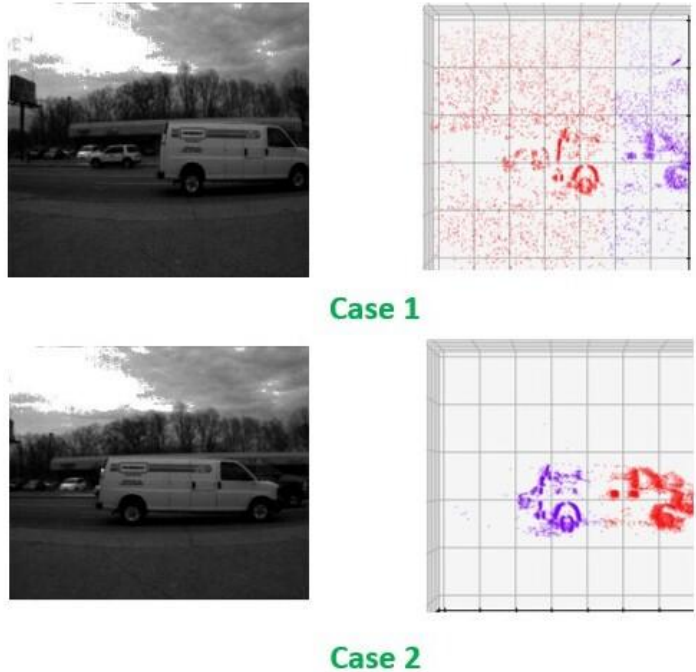

Fig. 4: Some failure cases. Case 1: During large noise in the dataset. Case 2: When one of the moving objects is larger than all the others.